\documentclass[runningheads]{llncs}

\usepackage{graphicx}
\usepackage{comment}
\usepackage{marvosym}
\usepackage{times}
\usepackage{amsmath}
\usepackage{latexsym}
\usepackage{microtype}

\usepackage{amssymb}
\usepackage{bm}
\usepackage{times}
\usepackage{graphicx}
\usepackage{multirow}
\usepackage{arydshln}
\usepackage{multicol}
\usepackage{enumerate}
\usepackage{booktabs}

\usepackage{color}
\usepackage{array}

\usepackage{enumitem}
\usepackage[T1]{fontenc}

\usepackage[utf8]{inputenc}

\usepackage{microtype}

\setitemize[1]{leftmargin=10pt,itemsep=0.5pt,partopsep=0pt,parsep=0.5pt,topsep=0pt}

\title{MedDG: An Entity-Centric Medical Consultation Dataset for Entity-Aware Medical Dialogue Generation}

\author{
Wenge Liu\inst{1}\thanks{W. Liu, J. Tang, and Y. Cheng—Equal contribution.}, Jianheng Tang\inst{2}$^{\star}$, Yi Cheng\inst{3}$^{\star}$, \\Wenjie Li$^{3}$, Yefeng Zheng\inst{4}, Xiaodan Liang\inst{1}\textsuperscript{(\Letter)}}
\institute{
Sun Yat-sen University
\and Hong Kong University of Science and Technology
\and Hong Kong Polytechnic University
\and Tencent Jarvis Lab\\
     \email{\{kzllwg,sqrt3tjh,xdliang328\}@gmail.com, alyssa.cheng@connect.polyu.hk, \\ cswjli@comp.polyu.edu.hk,       yefengzheng@tencent.com} }

\begin{document}
\maketitle
\begin{abstract}
Medical dialogue systems interact with patients to collect symptoms and provide treatment advice. 
In this task, medical entities (e.g., diseases, symptoms, and medicines) are the most central part of the dialogues. However, existing datasets either do not provide entity annotation or are too small in scale. 
In this paper, we present MedDG, an entity-centric medical dialogue dataset, where medical entities are annotated with the help of domain experts.
It consists of 17,864 Chinese dialogues, 385,951 utterances, and 217,205 entities, at least one magnitude larger than existing entity-annotated datasets. 
Based on MedDG, we conduct preliminary research on entity-aware medical dialogue generation by implementing several benchmark models.
Extensive experiments show that the entity-aware adaptions on the generation models consistently enhance the response quality but there still remains a large space of improvement for future research. 
The codes and the dataset are released at \url{https://github.com/lwgkzl/MedDG}.
\end{abstract}

\section{Introduction}

\begin{figure}[t]
\centering
  \includegraphics[width=.95\linewidth]{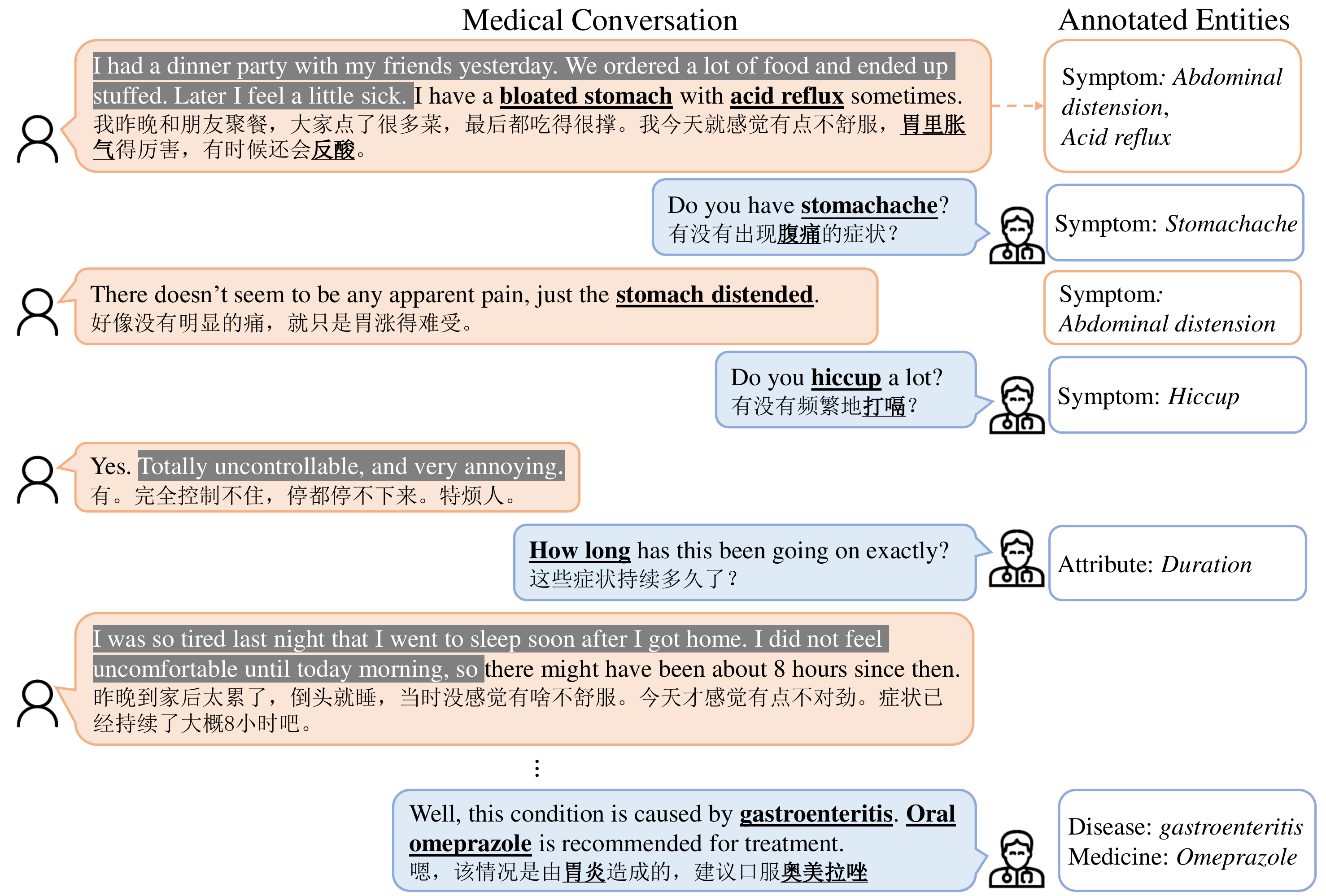}
\caption{An example medical consultation dialogue in the MedDG dataset. The entity-related text spans are underlined and their corresponding normalized entities are annotated in the right column. The gray part of the conversation are chit-chat content that barely contributes to medical diagnosis.}
\vspace{-4mm}
\label{fig:intro}
\end{figure}

Online medical consultations have been playing an increasingly important role. 
According to~\cite{mccall2020could}, telemedicine substantially rose from 10\% to 75\% of general medical consultations before and after the COVID-19 pandemic in UK.
However, patients often describe their situations verbosely and communicate with doctors inefficiently online. To address this dilemma, there has been growing research interest in developing automatic medical dialogue systems, which can assist doctors in pre-collecting symptoms and give patients preliminary treatment advice in time.

Different from generic dialogues, medical conversations are characterized by the following features. 
1) \textbf{Entity-Centric.} Medical entity terms are the most central part of medical conversations. They serve as the foundation of accurate diagnosis and therapy. 
2) \textbf{Redundancy.} Medical conversations usually contain redundant content that barely contributes to the medical diagnosis, as shown in the {gray part} of Fig. \ref{fig:intro}. 
Specifically, we sample 1,000 medical conversations from an online medical consultation platform, \emph{Doctor Chunyu} and manually label the content directly associated with symptom inquiry/description, disease diagnosis, examination, and therapy. 
We find that 40.1\% of the sentences posted by the patients are redundant chit-chat. 
3) \textbf{Changeable Inquiry Order.} In medical conversations, the doctor's symptom inquiry order is usually changeable. As in Fig. \ref{fig:intro}, the order of querying ``stomachache'' and ``hiccup'' can be exchanged. 

Given the above considerations, we argue that it would be beneficial to construct a medical conversation dataset, where the entity-related text spans and the corresponding normalized entities are annotated, as in Fig. \ref{fig:intro}. 
On the one hand, the annotation of entities could guide the introduction of external knowledge and help the medical dialogue system learn to focus on the central part of conversations \cite{liu2021heterogeneous}, contributing to a better understanding of the dialogue context and a more accurate diagnosis. 
On the other hand, it could facilitate a more comprehensive evaluation of medical dialogue generation by comparing the generated entity terms with the ones appearing in the following ground-truth responses.
Existing datasets, however, either do not provide such entity annotation~\cite{meddialog} or are too small in scale~\cite{DBLP:conf/emnlp/LinHCTWC19,mie,xu2019end,wei2018task}. 

In this paper, we present MedDG, an entity-centric Medical Dialogue dataset, with 17,864 Chinese dialogues, 385,951 utterances, and 217,205 entities. As demonstrated in Fig.~\ref{fig:intro}, in MedDG the entity-related text spans are first extracted from the original text and their corresponding normalized entities are then annotated. 
Specifically, we annotate five categories of entities, including disease, symptom, medicine, examination, and attribute. Table~\ref{tab:data_stat} lists the most frequent entities in each category in MedDG.
Compared with existing entity-annotated medical dialogue datasets, MedDG is one (or two) magnitude larger in terms of dialogue and utterance number and the entities covered in MedDG are diverse, as shown in Table~\ref{tab:data_comp}. Besides, the average and median occurrence times of each entity in MedDG are significantly higher than other corpora, demonstrating that models can learn features of each entity better with MedDG. 

\begin{table*}[tbp]
\centering
\caption{Comparison of entity-annotated medical dialogue datasets. The two columns below ``\# Entity Frequency'' refer to the average (Avg.) and median (Med.) occurrence time of an entity.}
\scalebox{0.85}{
\begin{tabular}{l|c|c|c|cc}
\specialrule{1.2pt}{1pt}{1pt}
\multirow{2}{*}{\textbf{Dataset}} & \multirow{2}{*}{\textbf{\# Dialogues}} &\multirow{2}{*}{\textbf{\# Utterances}} & \multirow{2}{*}{\textbf{\# Entity Labels}}& \multicolumn{2}{c}{\textbf {\# Entity Frequency}} \\ \cline{5-6}
 & & & & Avg. & Med.\\
\specialrule{1.2pt}{0pt}{1pt}
MZ~\cite{wei2018task}  & 710 & - & 70 & 67.06 & 33 \\
DX~\cite{xu2019end} & 527 & 2,816 & 46 & 65.95 & 12  \\
CMDD~\cite{lin-etal-2019-enhancing} & 2,067 & 87,005 & \textbf{161} & 194.39  & 18 \\
MIE~\cite{mie}  & 1,120 & 18,129  & 71 & 93.70 & 64 \\
\specialrule{0.7pt}{1pt}{1pt}
\textbf{MedDG}  &\textbf{17,864} & \textbf{385,951} & 160 & \textbf{1357.64} & \textbf{428} \\
\specialrule{1.2pt}{1pt}{0pt}
\end{tabular}}
\label{tab:data_comp}
\end{table*}
\begin{table*}[tbp]
\centering
\caption{Occurrence percentage of top four entities in each category.}
\scalebox{0.85}{
\begin{tabular}{lc|lc|lc|lc|lc}
\specialrule{1.2pt}{1pt}{1pt}
\multicolumn{2}{c|}{\textbf {Disease}} & \multicolumn{2}{c|}{\textbf {Symptom}} & \multicolumn{2}{c|}{\textbf {Medicine}} & \multicolumn{2}{c|}{\textbf {Examination}} & \multicolumn{2}{c}{\textbf {Attribute}} \\\specialrule{0.7pt}{1pt}{1pt}
Entity & \% & Entity & \%  & Entity & \% & Entity & \% & Entity & \%\\ 
\specialrule{1.2pt}{0pt}{1pt}
Gastritis  & 4.59 & Diarrhea  & 6.00 & Omeprazole  & 2.57 & Gastroscope  & 3.59 & Duration  & 3.77 \\
Enteritis  & 3.28 & Abdominal pain  & 5.71 & Motilium  & 1.01 & Colonoscopy
  & 2.12 & Risk factor  & 1.47 \\
Constipation  & 2.85 & Abdominal ext...  & 4.31 & Mosapride  & 0.95 & Stool test  & 0.46 & Pain degree  & 0.72 \\
Influenza  & 1.11  & Vomit  & 3.56 & Rabeprazole  & 0.80 & Blood test  & 0.39 & Body part  & 0.48 \\
\specialrule{0.7pt}{1pt}{1pt}
\textbf{Total} &13.70 & \textbf{Total} & 55.22& \textbf{Total} &16.76 & \textbf{Total} &7.89 & \textbf{Total} &6.42 \\
\specialrule{1.2pt}{1pt}{0pt}
\end{tabular}}
\label{tab:data_stat}
\end{table*}

Based on MedDG, we conduct preliminary research on entity-aware medical dialogue generation. 
Several benchmark models are implemented respectively for entity prediction and entity-aware generation. 
Experimental results show that our entity-aware adaptions of the generation models consistently enhance the response quality, demonstrating that the entity-aware methodology is a worth-exploring direction for this task. 
In summary, our contributions are three-fold: 
1) We collect MedDG, a large-scale entity-centric medical dialogue dataset, at least one magnitude larger than existing entity-annotated datasets; 
2) We conduct preliminary research on entity-aware medical dialogue generation by implementing multiple benchmark models and conducting extensive evaluation;
3) We conduct extensive analysis of the benchmark models and demonstrate that the entity-aware methodology consistently achieves better response quality.

\section{Related Work}

\subsubsection{Medical Dialogue Systems}

Tasks related to the understanding of medical dialogues have been researched a lot, such as information extraction~\cite{mie}, relation prediction~\cite{du-etal-2019-learning}, symptom prediction~\cite{lin-etal-2019-enhancing} and slot filling~\cite{scatter}. 
However, the generation part of medical dialogue systems was rarely investigated. 
Early attempts came from~\cite{early1,early2,liu2016augmented}. More recently, ~\cite{xu2019end,liu2022my} introduced an end-to-end medical dialogue system based on pre-defined templates, which suffers from the problem of inflexibility.
\cite{meddialog} took an initial step in neural-based medical dialogue generation. They pre-trained several dialogue generation models on large-scale medical corpora and studied the transferability of models to the low-resource COVID-19 dialogue generation task. However, they did not consider medical entities, which are the most central part of medical dialogues.

\subsubsection{Medical Dialogue Datasets}
\cite{wei2018task} first launched a dataset for medical diagnosis, but it only contains some structured data related to the consultation content, rather than dialogues in the form of natural language. The DX dataset~\cite{xu2019end} contains 527 medical dialogues, but the utterances are automatically generated with templates. ~\cite{lin-etal-2019-enhancing} collected the CMDD dataset with 2,067 dialogues on four pediatric diseases, and ~\cite{mie} released the MIE dataset with 1,120 dialogues on six cardiovascular diseases. However, both CMDD and MIE were proposed for dialogue understanding tasks, not large enough to build a competitive dialogue system. 
More recently, ~\cite{meddialog} collected 3.4 million Chinese medical dialogues  from the Internet that make up an extra-large dataset, but the dialogues in it are all unlabeled, so its quality is not guaranteed by human-annotators and it is unable to facilitate research on entity-aware medical dialogue generation. 
Table~\ref{tab:data_comp} summarizes the statistics of public entity-annotated Chinese medical dialogue corpora. Among them, our MedDG dataset is the largest and includes diverse entity types.
\section{MedDG Dataset}

\subsection{Data Collection}\label{sec:dataCollect}
The dialogues in MedDG are collected from the gastroenterology department of an online Chinese medical consultation website, \emph{Doctor Chunyu},\footnote{https://www.chunyuyisheng.com/} where patients can submit posts about their problems and consult qualified doctors for professional advice. 
In total, more than 100,000 dialogues are collected. 
Then some low-quality dialogues are filtered out if the number of utterances is not between 10 and 50, or if it contains nontext information, such as graphics or sound.
After filtering, there remain 17,864 dialogues, with 385,951 utterances. 
We select consultations in the gastroenterology department as their related inquiries are common and diverse. In addition, preliminary treatment advice for gastrointestinal problems relies less on medical examinations than the other diseases, and they are thus more suitable for online consultations. 

To avoid privacy issues, the \emph{Doctor Chunyu} website has already filtered the sensitive information when releasing consultation content. 
After collecting the online data, we further ensure that there remains no private information by involving two checking approaches.
First, we design a set of regular expression rules to extract private information, like ages, telephone numbers, and addresses, and then manually check the extracted results. 
Second, the annotators are asked to report to us if they observe any privacy issue during entity annotation. 
Neither of the above approaches find any sensitive information, proving that the source data has a very low risk of privacy issues.

\subsection{Entity Annotation}\label{sec:autoLabel}
We then conduct entity annotation on the collected dialogues in the form similar to \cite{DBLP:conf/emnlp/LinHCTWC19}: the entity-related text spans are first extracted from the original text and then normalized into our pre-defined normalized entities, as shown in Fig. \ref{fig:intro}.
To control costs within a reasonable range, we design a semi-automated annotation pipeline. First, we determine the entities to be included in the annotation. Then, domain experts manually annotate 1,000 dialogues in the dataset. Finally, an automatic annotation program is implemented based on the human annotation to label the remaining data. 

\subsubsection{Entity Determination}
After discussing with domain experts, we choose five categories of entities for annotation: disease, symptom, examination and attribute. 
Then, we further define a total of 160 entities to be included in these categories, referring to the terminology lists in the Chinese medical knowledge graph CMeKG \cite{CMeKG} and their frequency in the dataset. 
There are 12/62/62/20/4 entities, respectively, in the disease/symptom/medicine/examination/attribute category.
Table~\ref{tab:data_stat} lists the four most frequent entities in each category and their frequency distribution.

\subsubsection{Human Annotation} 
A total of 1,000 dialogues in the dataset are manually annotated. 
Eight annotators with more than one year of consultation experience are involved in the annotation process. They are asked to spend an average of 5 minutes on each dialogue, and paid with \$50 per hour. 
To guide the annotators, we write the initial version of annotation manual through discussions with domain experts. 
The annotators are asked to first conduct trial annotation based on this manual, and report their problems on it. 
Then we revise the manual based on the found issues, and launch the formal annotation process. 
Each dialogue is labeled by two annotators independently and the inconsistent part is further judged by a third annotator. 
The Cohen's kappa coefficient between different pairs of annotators is between 0.948 and 0.976, indicating a strong agreement between annotators (1.0 denotes complete agreement).

\subsubsection{Semi-automatic Annotation}
We then develop an automatic annotation program to label the remaining data. 
As its initial version, a set of regular expression rules are first designed based on the 1,000 human-labeled dialogues. That is, they can accurately cover all the entity annotation in the manually-annotated data.
Then, to ensure its annotation quality, we randomly pick 2,000 utterances and invite experts to manually label them again. By comparing the annotation from the experts and the automatic program, we evaluate the annotation quality and further improve the annotation program repeatedly. 
The above revision procedure is repeated four times. 
The final version of the annotation program achieves 96.75\% annotation accuracy on the 2,000 manually evaluated utterances, which supports the hypothesis that annotation accuracy is greater than 95\% on the whole dataset ($p$-value < 0.01).

\subsection{Dataset Statistics}
\begin{table}[t]
\centering
\caption{Statistics about the number of dialogues, utterances, tokens, and entities in MedDG.}
\begin{tabular}{l|c}
\specialrule{1.2pt}{1pt}{1pt}
\# Dialogues  & 17,864\\
\# Utterances & 385,949\\
\# Tokens & 6,829,562\\
\# Entities & 217,205\\ 
\specialrule{1.0pt}{0pt}{0pt}
Avg. \# of utterances in a dialogue & 21.64\\
Avg. \# of entities in a dialogue & 12.16\\
Avg. \# of tokens in an utterance & 17.70\\
Avg. \# of entities in an utterance & 0.56\\
\specialrule{1.2pt}{1pt}{1pt}
\end{tabular}
\vspace{-3mm}
\label{tab:stat}
\end{table}
There are 14,864/2,000/1,000 dialogues in the train/dev/test sets. 
The human-annotated dialogues are put in the test set and the rest are randomly divided into the training and the development sets. 
Table~\ref{tab:stat} presents statistics about the number of dialogues, utterances, tokens, and entities in MedDG. %
Table \ref{tab:data_stat} illustrates the distribution of categories and top four entities in each category. 
We can see that the proportion of symptoms is the highest, accounting for 55\% of the total entity occurences. 

\section{Experiments}
Based on MedDG, we conduct prelimiary research on entity-aware medical dialogue generation. 
Specifically, we investigate several pipeline methods that first predicts the entities to be mentioned in the upcoming response and then generates the response based on the prediction results. 
The formal definition of the two tasks involved (\emph{entity prediction} and \emph{entity-based response generation}) are as below.
Formally, at each dialogue turn of the consultation system, it has access to the dialogue history consisting of a sequence of utterances from either the patient or the system itself, $X_{1:K}=\{X_1,X_2,...,X_K\}$. 
The task of \emph{entity prediction} is to predict the set of entities $ E_{K+1}=\{e_{K+1}^1,e_{K+1}^2,...,e_{K+1}^l\}$ to be mentioned in the upcoming response $X_{K+1}$.
As for \emph{entity-based response generation}, it is to generate the response $X_{K+1}$, based on the dialogue history $X_{1:K}$ and the entity prediction results $E'_{K+1}$.

\subsection{Baselines}
\subsubsection{Entity Prediction}\label{sec:entPred}
Entity prediction is done for each pre-defined entity $e_i$ to decide whether it should be included in the upcoming response. 
We use first neural encoder to obtain the context vector of the dialogue history, which is then passed to a fully connected layer with a sigmoid function.
Formally, the entity prediction process is defined as:
\begin{equation}
\begin{aligned}
h_D &= \text{Encoder}(X_{1:K})\\
g(e_i) &= \text{Sigmoid}(w_p^Th_D)
\end{aligned}
\end{equation}
where $h_D$ is the context vector of the dialogue history and $w_p$ is a learnable parameter.
We use a binary cross entropy loss computed as:
\begin{equation}
L_e = -y_i \text{log}(g(e_i))-(1-y_i) \text{log}(1-g(e_i))
\end{equation}
where $y_i \in \{0, 1\} $ is the ground truth indicating whether $e_i$ will be included in the upcoming response or not.
The following baseline models are implemented to encode the dialogue history: 
\textbf{LSTM}~\cite{hochreiter1997long}, 
\textbf{TextCNN}~\cite{kim-2014-convolutional}, 
\textbf{BERT-wwm}~\cite{cui2019pre}, 
\textbf{PCL-MedBERT}~\cite{PCL-MedBERT}, 
\textbf{MedDGBERT}. MedDGBERT is a BERT-based model, pretrained on the training set of MedDG with a designed entity-prediction task, which is to predict the masked entities mentioned in the dialogue.

\subsubsection{Entity-based Response Generation}\label{sec:entResponse}
For entity-based response generation, we implement the following two approaches:

\paragraph{Retrieval.} Given the predicted entities $E'_{K+1}$, the retrieval-based method chooses a doctor's utterance $U_{R}$ in the training set of MedDG as its response. $U_{R}$ is randomly picked from the utterances which satisfy the condition that their annotated entities $E_{R}$ are the minimum coverage of all the predicted entities $E'_{K+1}$.

\paragraph{Entity Concatenation.} 
Given the predicted entities $E_{K+1}$, we concatenate them with the dialogue history $X_{1:K}$ and then pass them to a generation model. The averaged negative log-likelihood of the target sequence $X_{K+1}=\{w_{K+1}^1, w_{K+1}^2, ..., w_{K+1}^T\}$ is used as the generation loss:
\begin{equation}
    \mathcal L_g=-\frac{1}{T}\sum_{t=1}^T \log P(w_{K+1}^t).
\end{equation}
We compare the following baseline models to generate responses with entity concatenation:
\textbf{Seq2Seq}~\cite{seq2seq}, 
\textbf{HRED}~\cite{serban2016building}, 
\textbf{GPT-2}~\cite{gpt}, 
\textbf{DialoGPT}~\cite{Zhang2020DialoGPTLG}, 
\textbf{BERT-GPT}~\cite{meddialog}, and 
\textbf{MedDGBERT-GPT}. BERT-GPT combines the BERT encoder and GPT decoder and is fine-tuned on the MedDialog dataset \cite{meddialog}. MedDGBERT-GPT has the same architecture as BERT-GPT, but its parameter of the BERT encoder are replaced with the weights in MedDGBERT, one of our entity prediction baselines.

\begin{table*}[t!]
\caption{Results of the entity prediction task on MedDG, in terms of the precision, recall, F1 scores of all the entities, and the F1 scores of each category. Note that all metrics are presented in percentage (\%).}
\centering
\begin{tabular}{l|ccc|ccccc}
\specialrule{1.2pt}{1pt}{1pt}
\textbf{Model}  &  ~\textbf{P}~   & ~\textbf{R}~  & ~\textbf{F1}~ & ~\textbf{F1}$_\text{\textbf{D}}$~ &  ~\textbf{F1}$_\text{\textbf{S}}$~  &  ~\textbf{F1}$_\text{\textbf{M}}$~   & ~\textbf{F1}$_\text{\textbf{E}}$~  &  ~\textbf{F1}$_\text{\textbf{A}}$~   \\
\specialrule{1.2pt}{1pt}{1pt}
LSTM & 25.34 & 27.75 &26.49 & 31.18 & 21.72 & 15.66 & 25.05  & 48.95  \\
TextCNN  & 22.37 &30.12 &25.67 & 29.54 & 20.55  & 19.01 & 23.58  & 50.33 \\
BERT-wwm & 26.05 &31.09 &28.35 & 31.66 & 24.27 & 19.82 & 26.03 & \textbf{52.44}  \\
PCL-MedBERT & \textbf{26.46}  & 33.07 & 29.40 & \textbf{33.72} & 25.62 & 20.78  & 27.49 & 46.85\\
MedDGBERT &25.34 &\textbf{36.20}  &\textbf{29.81}  & 33.29 &\textbf{26.39}  &\textbf{21.35} &\textbf{27.60} & 49.41\\
\specialrule{1.2pt}{1pt}{1pt}
\end{tabular}
\vspace{-3mm}
\label{tab:result1}
\end{table*}
\subsection{Implementation Details}
For the entity prediction methods, prediction is conducted for each predefined entity as a binary classification task, and the final predicted entity set is made up of the ones whose predicted probabilities are greater than 0.5. 
For the entity-based response generation methods, the entities concatenated in the input of the response generation models are automatically predicted, with the strong entity prediction baseline model MedDGBERT; and the dialogue history is abbreviated to the last patient's utterance to speed up the experiments.
For the three LSTM-based models (LSTM, Seq2seq, HRED), we implement a single-layer LSTM as the encoder or decoder. 
The dimensions of the word embedding and the hidden states in LSTM are set to 300. We use the Adam optimizer with a mini-batch size of 16 and set the initial learning rate to 0.001. 
For the other baseline models, we directly follow the implementations in their original papers. 
The number of training epoch is all bounded at 30, and training will be early-stopped after 5 epochs without improvement.

\subsection{Results on Entity Prediction}
For the entity prediction task, the evaluation metrics include precision, recall and F1. In addition to the scores evaluated on all the entities, we also calculate the F1 score of the ones in each entity categories, denoted as F1$_\text{D}$ (Disease), F1$_\text{S}$ (Symptom), F1$_\text{M}$ (Medicine), F1$_\text{E}$ (Examination), and F1$_\text{A}$ (Attribute). The results are shown in Table~\ref{tab:result1}.

Compared with classic RNN and CNN encoders, we can see that the three BERT-based encoders (BERT-wwm, PCL-MedBERT, and MedDGBERT) consistently achieve better performance than LSTM and TextCNN in terms of all metrics. 
MedDGBERT outperforms the other two BERT variants in most of the metrics, which demonstrates the effectiveness of including entity prediction in the pretraining process to integrate medical entity knowledge. 
Comparing the F1 scores of each category, we find that the performance on the attribute category is significantly superior to the one on the others. 
It is probably because entities in the attribute category follow relatively fixed patterns. For example, after patients’ description of their symptoms, doctors would usually further ask about attributes of the symptoms, like duration and pain degree. 
In comparison, entity prediction on the other four categories are much more challenging, as they contain more types of entities and more heavily rely on domain knowledge. The diversity of different doctors' diagnosis strategies also increases the difficulty.

\subsection{Results on Response Generation}

\begin{table*}[t!]
\caption{Results of the response generation task on MedDG. ``w/o Ent.'' refers to the entity-ablated version of the above method, which does not perform entity prediction and directly take the dialogue history as input. P$_\text{Ent}$, R$_\text{Ent}$, F1$_\text{Ent}$ stand for precision, recall and F1 of the entities mentioned in the generated results. Note that all metrics are normalized to $[0,100]$.}
\centering
\begin{tabular}{l|cccc|ccc}
\specialrule{1.2pt}{1pt}{1pt}
\textbf{Model}  & \textbf{BLEU-1} &\textbf{BLEU-4} & \textbf{Distinct-1}  & \textbf{Distinct-2} &~\textbf{P}$_\text{\textbf{Ent}}$~& ~\textbf{R}$_\text{\textbf{Ent}}$~ & ~\textbf{F1}$_\text{\textbf{Ent}}$~\\
\specialrule{1.2pt}{1pt}{1pt}
Retrieval & 23.08 & 12.58 & 0.62 & 9.98 & 11.44 & \textbf{36.25} & 17.39 \\\cdashline{1-8}[0.8pt/2pt]
Seq2Seq & 35.24 & 19.20 & 0.75 & 5.32 & 12.41 & 25.65 & 16.73  \\
$\quad$w/o Ent. & 26.12 & 14.21 & 0.88 & 4.77 & 14.07 & 11.45 & 12.63 \\
\cdashline{1-8}[0.8pt/2pt]
HRED  & \textbf{38.66} & 21.19 & 0.75 & 7.06 & 12.01  & 26.78 & 16.58\\
$\quad$w/o Ent. & 31.56 & 17.28 & 1.07 & 8.43 & 13.29 & 11.25  & 12.18\\
\cdashline{1-8}[0.8pt/2pt]
GPT-2  & 30.87 & 16.56 & 0.87 & 11.20 & \textbf{14.51} & 20.76   & 17.08\\
$\quad$w/o Ent.  & 29.35 & 14.47 & \textbf{1.26} & 13.53 & 7.33  & 12.22 & 9.17\\
\cdashline{1-8}[0.8pt/2pt]
DialoGPT  & 34.90 & 18.61 & 0.77 & 9.87 &13.53  &   21.16& 16.51\\
$\quad$w/o Ent.  & 34.57 & 18.09 & 0.50 & 9.92 & 11.30  & 9.99 & 10.61\\
\cdashline{1-8}[0.8pt/2pt]
BERT-GPT &36.54 &23.84 &0.65 &11.25 &12.74 &28.71 &17.65\\
$\quad$w/o Ent. & 32.69 & 21.18 & 0.94 & 13.80 & 10.13 & 11.13 & 10.61 \\ \cdashline{1-8}[0.8pt/2pt]
MedDGBERT-GPT   &36.62 & \textbf{23.99}  & 0.63  & 11.04 & 13.78 & 28.55 & \textbf{18.59}\\
$\quad$w/o Ent.  &32.62 &20.95 &0.93 & \textbf{13.88} &10.89 &11.59 &11.23 \\
\specialrule{1.2pt}{1pt}{1pt}
\end{tabular}
\label{tab:result2}
\end{table*}
\begin{table}[t!]
\caption{Human evaluation results of different response generation methods in terms of fluency (Flu.), relevance (Rel.) and expertise (Exp.) on a scale of [1-5]. }
\centering
\begin{tabular}{l|ccc}
\specialrule{1.2pt}{1pt}{1pt}
\textbf{Model} & ~\textbf{Flu.}~ &  ~\textbf{Rel.}~ & ~\textbf{Exp.}~ \\
\specialrule{1.2pt}{1pt}{1pt}
Retrieval &\textbf{4.23} &\textbf{4.18} &\textbf{4.29}\\\cdashline{1-4}[0.8pt/2pt]
HRED  &3.46 &3.16 &3.12 \\
$\quad$w/o Ent. &2.84 &2.73 &2.40\\ 
\cdashline{1-4}[0.8pt/2pt]
MedDGBERT-GPT &3.93 &3.62 &3.55\\
$\quad$w/o Ent. &3.27 &2.83 &2.69\\
\specialrule{1.2pt}{1pt}{1pt}
\end{tabular}
\label{tab:human}
\end{table}

\subsubsection{Automatic Evaluation}
For automatic evaluation, we assess the generated response in two aspects: the text quality and the medical entities it covers.
To evaluate text quality, we use the \textbf{BLEU-1}, \textbf{BLEU-4}~\cite{chen2014systematic}, \textbf{Distinct-1}, and \textbf{Distinct-2}~\cite{li2016diversity} metrics. BLEU-1/4 measures lexical similarity, while Distinct-1/2 evaluates text diversity. 
To evaluate the medical entities covered in the response, we first use the automatic annotation program described in Sec. \ref{sec:autoLabel} to extract the entities mentioned in the generated responses, and then compare them with the gold entities in terms of precision, recall, and F1, named as ~\textbf{P}$_\text{\textbf{Ent}}$, ~\textbf{R}$_\text{\textbf{Ent}}$, ~\textbf{F1}$_\text{\textbf{Ent}}$, respectively.  
The results are presented in Table~\ref{tab:result2}. The lines begins with ``\textbf{w/o Ent}'' show the ablation results of the corresponding models, which do not perform entity prediction and directly take the dialogue history as input. By analyzing Table~\ref{tab:result2}, we can obtain the following insights. 

\paragraph{Entity-aware adaptions consistently improve the generation quality.} Comparing the entity-aware methods and their entity-ablated versions (denoted as ``w/o Ent.'' in Table \ref{tab:result2}), we can see that the entity-aware adaptions significantly enhance both the BLEU  and entity-related scores, demontrating the effectiveness of entity-aware methods in improving generation quality. For instance, MedDGBERT-GPT has $14.51\%$ relative improvement on BLEU-4 and $65.54\%$ on F1$_{\text{Ent}}$ compared with its entity-ablated version. 
The entity prediction results utilized by the entity-aware methods are provided by MedDGBERT, whose F1 score is no more than 30\%, but the improvement is still significant.
More improvement could be expected with stronger entity prediction models in the future.
Besides, Distinct-1/2 drop after entity-aware adaptions in some cases, it is mainly because the entity-aware methods tend to generate more specific content, following relatively fixed patterns.

\paragraph{Retrieval vs. Generation. }
As our retrieval-based method directly respond with utterances in the training set based on the predicted entities without consideration of the contextual information, its BLEU score is lower than generative models, while the entity-related metrics are relatively high. 
Notably, the entity recall is significantly higher than all the generative methods, because our retrieval strategy ensures that all the predicted entities are covered in the response.

\paragraph{LSTM-based vs. GPT-based.} 
Though with much fewer parameters, the BLEU scores of HRED are not lower than those of most GPT-based models. One possible reason is that the GPT models are pre-trained mainly on large-scale chit-chat data, so they tend to generate responses irrelevant to the medical domain. 
It also explains why GPT-based models have higher distinct scores. 

\begin{table*}[t!]
\caption{Examples of the generated responses from different methods. The left column lists the patient's utterances, gold responses, predicted and gold entities. The three presented turns belong to the same dialogue.  The \underline{underlined} entities are the ones correctly predicted. }
\centering
\small
  \includegraphics[width=1\linewidth]{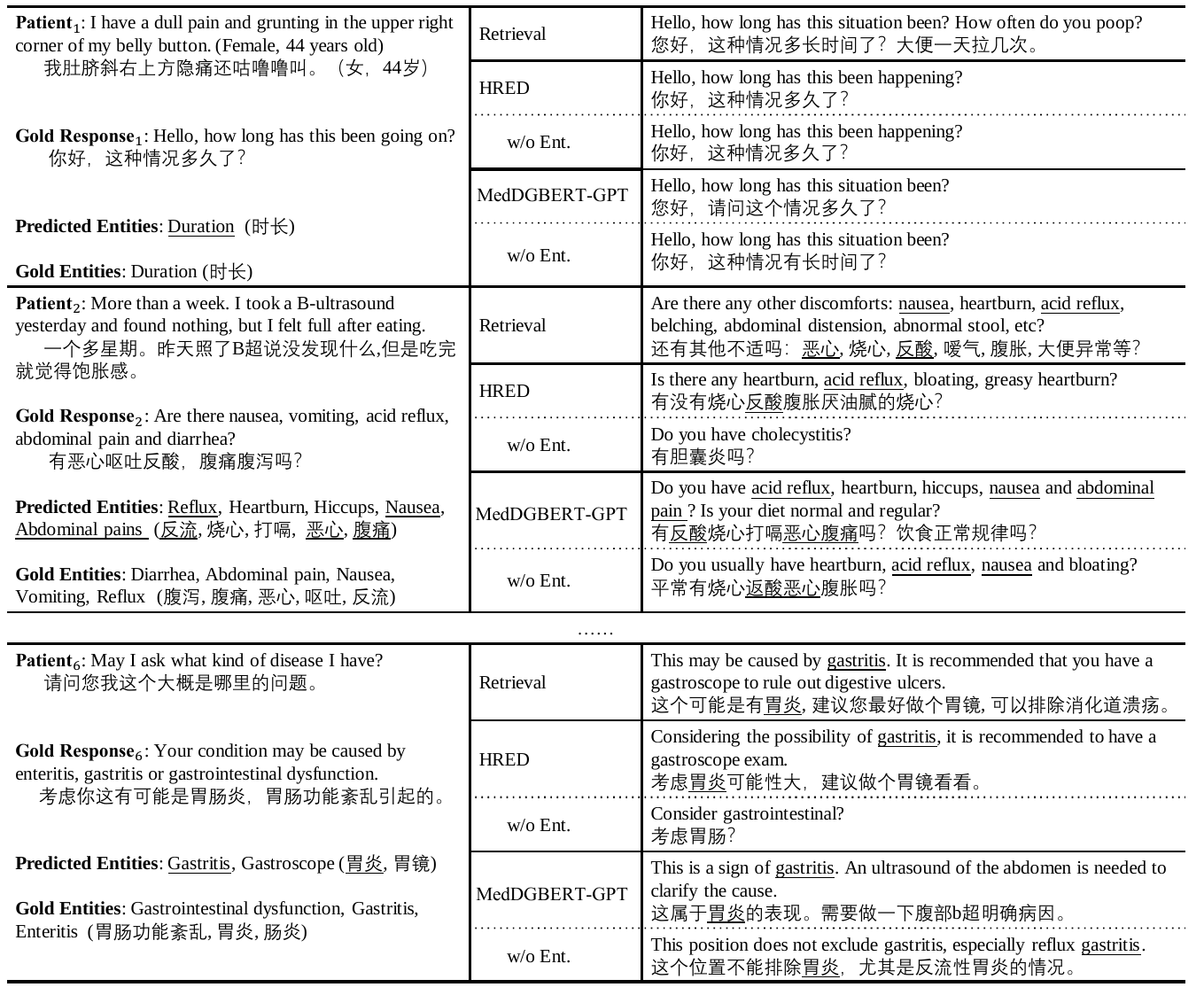}
\label{fig:case}
\vspace{-5mm}
\end{table*}

\subsubsection{Human Evaluation}  
We further conduct human evaluation of five selected generation methods, including Retrieval, HRED, MedDGBERT-GPT, and the entity-ablated versions of the latter two. 
We randomly pick 100 test dialogues, each with an average 9.71 utterances from the doctor.
Three annotators involved in our dataset annotation process are asked to rate these models' responses independently between 1 (poor) and 5 (good) in the following three dimensions. 
1) \textbf{Fluency}: It checks whether the response follows the grammar and semantically correct; 
2) \textbf{Expertise}: It checks whether the response content is appropriate from a professional medical perspective; 
3) \textbf{Relevance}: It checks whether the response is relevant to the dialogue history.

The results are shown in Table~\ref{tab:human}. Surprisingly, the retrieval-based method is significantly superior to other generative models on all metrics, especially in terms of sentence fluency, which is opposite to our automatic evaluation results. This is mainly because the fluency and quality of the retrieved response are guaranteed as they are all doctor-written sentences, while the BLEU metric only measure the similarity between the generation results and the reference utterances in terms of $n$-grams. 
Comparison of the four generative models can lead to the same conclusions as in the automatic evaluation that entity-aware methods can improve generation quality. The average Cohen's kappa scores between annotators are 0.41, 0.52, and 0.59 in fluency, relevance, and expertise, respectively.

\subsubsection{Case Study}
Table \ref{fig:case} gives three examples of the generated responses from different generaion methods. 
We can see that with the guidance of these predicted entities, the model responses are more informative and contain more correct entities. For example, in the second turn, HRED accurately includes the entity ``acid reflux'' under the guidance of predicted entities, while its entity-ablated version generates the wrong entity ``cholecystitis''. Compared with other methods, MedDGBERT-GPT generates the most number of correct entities in three turns. 

\section{Conclusion and Future Work}
In this paper, we proposed MedDG, a large-scale Chinese medical consultation dataset with annotation of rich medical entities. Based on MedDG, we conducted preliminary research on entity-aware medical dialogue generation, by implementing a pipeline method that first predicts entities to be mentioned and then conducts generation based on these entities. Through extensive experiments, we demonstrated the importance of taking medical entities into consideration. In the future, we will further investigate how to introduce domain knowledge to this task. One possible approach is to model the relationship between different medical entities in MedDG based on medical knowledge graphs like CMeKG \cite{CMeKG}.

\bibliography{anthology,custom}
\bibliographystyle{unsrt}
\appendix

\end{document}